\documentclass[conference]{IEEEtran}
\IEEEoverridecommandlockouts
\usepackage{cite}
\usepackage{amsmath,amssymb,amsfonts}
\usepackage{algorithmic}
\usepackage{graphicx}
\usepackage{textcomp}
\usepackage{xcolor}
\usepackage{tabularx}
\usepackage{booktabs}
\usepackage{multirow}
\usepackage{subcaption}
\def\BibTeX{{\rm B\kern-.05em{\sc i\kern-.025em b}\kern-.08em
    T\kern-.1667em\lower.7ex\hbox{E}\kern-.125emX}}
\begin{document}

% \title{Text-Aware Panoptic Symbol Spotting \\ in CAD Drawings 
\title{Text-Enhanced Panoptic Symbol Spotting \\ in CAD Drawings
%{\footnotesize \textsuperscript{*}Note: Sub-titles are not captured for https://ieeexplore.ieee.org  and
}

\author{
\IEEEauthorblockN{
Xianlin Liu\IEEEauthorrefmark{1},
Yan Gong\IEEEauthorrefmark{2},
Bohao Li\IEEEauthorrefmark{2},
Jiajing Huang\IEEEauthorrefmark{2},
Bowen Du\IEEEauthorrefmark{3},
Junchen Ye\IEEEauthorrefmark{3}\textsuperscript{*},
Liyan Xu\IEEEauthorrefmark{3}
}
\IEEEauthorblockA{
\IEEEauthorrefmark{1}\textit{Railway Siyuan Survey and Design Group Co., Ltd.}, Wuhan, China \\
1429737786@qq.com
}
\IEEEauthorblockA{
\IEEEauthorrefmark{2}\textit{CCSE Lab, Beihang University}, Beijing, China \\
\{gongy, libh, jiajinghuang\}@buaa.edu.cn
}
\IEEEauthorblockA{
\IEEEauthorrefmark{3}\textit{School of Transportation Science and Engineering, Beihang University}, Beijing, China \\
\{dubowen, junchenye, xuliyan\}@buaa.edu.cn
}
}

% \author{\IEEEauthorblockN{Xianlin Liu}
% \IEEEauthorblockA{\textit{Railway Siyuan Survey and Design Group Co., Ltd.} \\
% %\textit{name of organization (of Aff.)}\\
% Wuhan, China \\
% 1429737786@qq.com}
% \and
% \IEEEauthorblockN{Yan Gong}
% \IEEEauthorblockA{\textit{CCSE Lab, Beihang University} \\
% %\textit{name of organization (of Aff.)}\\
% Beijing, China \\
% \and
% \IEEEauthorblockN{Bohao Li}
% \IEEEauthorblockA{\textit{CCSE Lab, Beihang University} \\
% %\textit{name of organization (of Aff.)}\\
% Beijing, China  \\
% \and
% \IEEEauthorblockN{Jiajing Huang}
% %\textit{name of organization (of Aff.)}\\
% Beijing, China \\
% \and
% \IEEEauthorblockN{Bowen Du, Junchen Ye, Liyan Xu}
% \IEEEauthorblockA{\textit{School of Transportation Science and Engineering, Beihang University} \\
% %\textit{name of organization (of Aff.)}\\
% Beijing, China \\
% \email{\{dubowen,junchenye,xuliyan\}@buaa.edu.cn}}
% \and
% \IEEEauthorblockN{Junchen Ye}
% Beijing, China  \\
% \and
% \IEEEauthorblockN{Liyan Xu}
% Beijing, China  \\

%\author{
%\IEEEauthorblockN{Anonymous Authors}
%}

\maketitle

\begin{abstract}
With the widespread adoption of Computer-Aided Design(CAD) drawings in engineering, architecture, and industrial design, the ability to accurately interpret and analyze these drawings has become increasingly critical. Among various subtasks, panoptic symbol spotting plays a vital role in enabling downstream applications such as CAD automation and design retrieval.
Existing methods primarily focus on geometric primitives within the CAD drawings to address this task, but they face following major problems: they usually overlook the rich textual annotations present in CAD drawings and they lack explicit modeling of relationships among primitives, resulting in incomprehensive understanding of the holistic drawings. To fill this gap, we propose a panoptic symbol spotting framework that incorporates textual annotations. 
The framework constructs unified representations by jointly modeling geometric and textual primitives. Then, using visual features extract by pretrained CNN as the initial representations, a Transformer-based backbone is employed, enhanced with a type-aware attention mechanism to explicitly model the different types of spatial dependencies between various primitives. Extensive experiments on the real-world dataset demonstrate that the proposed method outperforms existing approaches on symbol spotting tasks involving textual annotations, and exhibits superior robustness when applied to complex CAD drawings.
\end{abstract}


\section{Introduction}
\begin{figure}[htbp]
    \centering
    \begin{subfigure}[b]{0.45\linewidth}
        \centering
        \includegraphics[width=\linewidth]{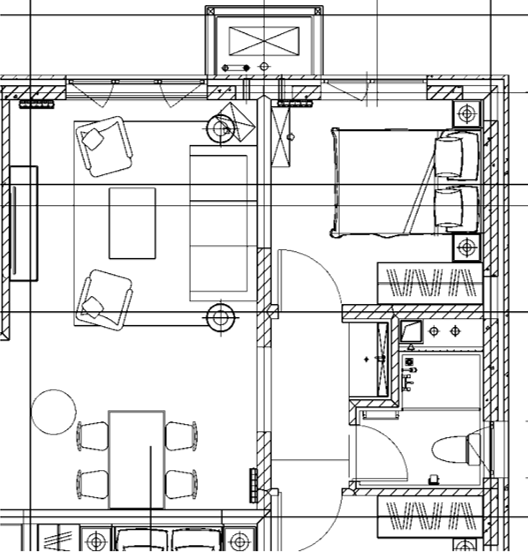}
        \caption{CAD drawings contain geometric symbols}
        \label{fig:sub-a}
    \end{subfigure}
    \quad % 比 \hfill 紧凑
    \begin{subfigure}[b]{0.45\linewidth}
        \centering
        \includegraphics[width=\linewidth]{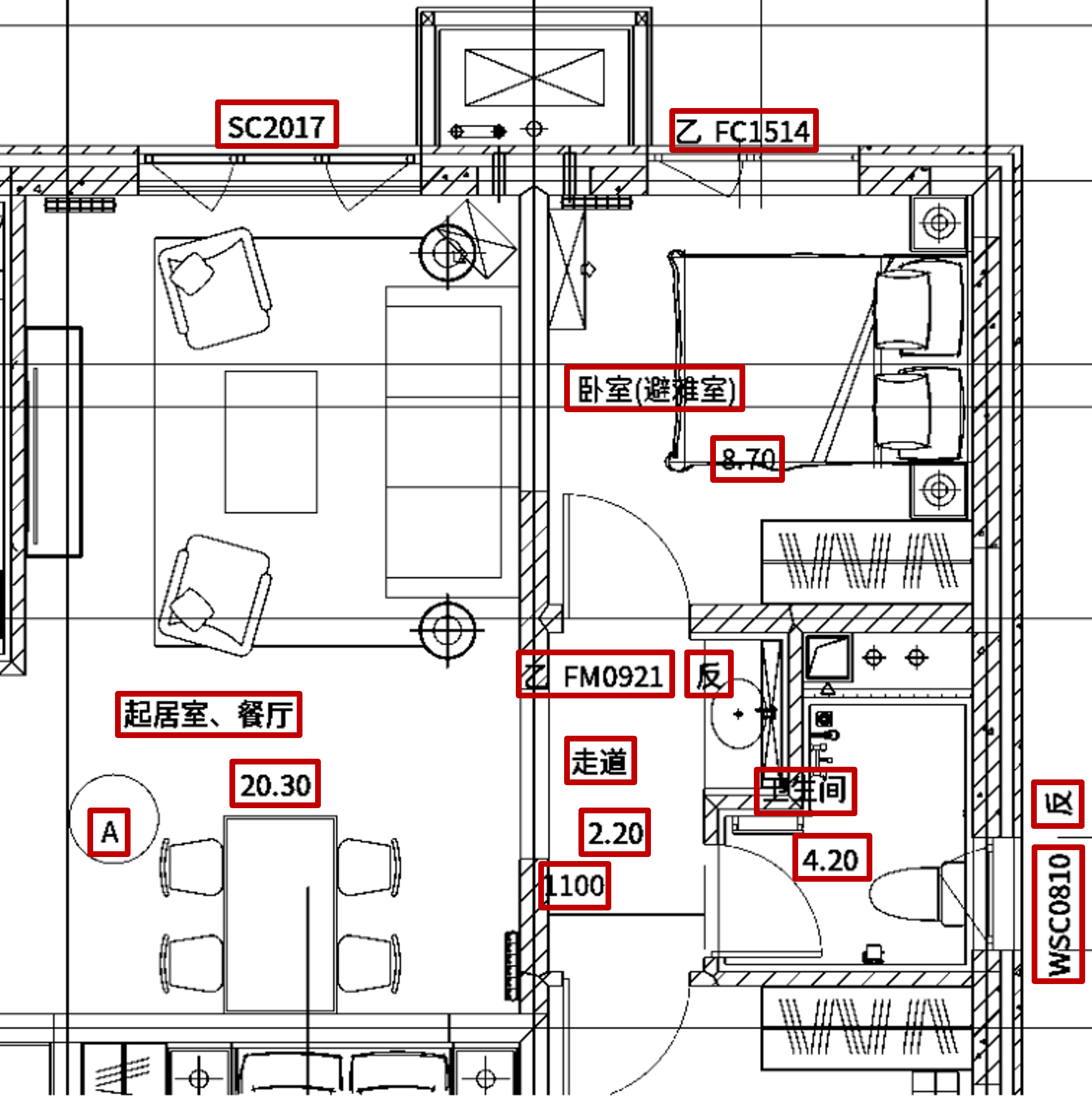}
        \caption{CAD drawings contain text annotations}
        \label{fig:sub-b}
    \end{subfigure}
    \caption{A sample of CAD drawings}
    \label{fig1}
\end{figure}

Computer-Aided Design (CAD) involves the use of computer technology to assist in the development of design models by generating precise 2D and 3D visual representations, which is widely adopted in real-world industries in architecture, engineering and construction (AEC) industries (e.g., mechanical manufacturing, architecture, electronics, and aerospace engineering)\cite{shivegowda2022review,hirz2017future,aouad2013computer,liu2019review}. Rather than relying on rasterized pixel data, CAD drawings are vector-based and constructed using fundamental geometric symbols such as lines, arcs, circles, and ellipses, which convey not only detailed structural configurations but also semantic information\cite{fan2021floorplancad}, as illustrated in Fig.~\ref{fig:sub-a}. CAD Symbol spotting refers to the process of detecting and recognizing these graphic symbols within a drawing\cite{rezvanifar2019symbol}. It serves as a foundational step toward semantic understanding of CAD drawings and is critical for numerous industrial applications, including intelligent design interpretation, automated modeling, and drawing retrieval\cite{rusinol2010symbol}.

Early studies on symbol recognition primarily focused on instance-level detection and classification of countable things, such as sofas, chairs, and beds \cite{rezvanifar2019symbol}. However, these approaches neglected the semantic understanding of uncountable stuff, which is equally critical in CAD environments. For example, the wall, which was typically represented by sets of parallel lines in CAD drawings \cite{fan2021floorplancad}, was frequently treated as background without proper semantic labeling \cite{rezvanifar2020symbol,nguyen2008symbol}. To address this limitation, Fan \cite{fan2021floorplancad}, inspired by the ideas in \cite{kirillov2019panoptic}, proposed the panoptic symbol spotting task, which unifies instance-level symbol detection with semantic recognition of uncountable structures. To tackle this task, some methods adopts the pixel-based strategy which integrates object detection and segmentation techniques \cite{pang2024pixel}, while others treats geometric primitives in CAD drawings as the basic processing units and employs transformer-based architectures \cite{fan2022cadtransformer} or graph-based approaches \cite{fan2021floorplancad,zheng2022gat} to model the primitives information.

Despite the impressive progress of these approaches, there remain two critical problems when applied to large-scale and complex real-world CAD floor plans:
1) \textbf{\textit{Lack of utilization of textual elements in CAD drawings.}}   
In practical CAD applications, drawings often consist not only of graphical elements constructed from geometric primitives but also a wide range of textual annotations, including dimension labels, symbol names, and functional descriptions, as illustrated in Fig.~\ref{fig:sub-b}. These textual elements usually provide semantic cues that complement the meaning of surrounding structures, serving as an essential source for understanding design intent\cite{li2013symbol}. However, current methods mainly focus on geometric primitives, overlooking the semantic and contextual information embedded in textual content\cite{fan2021floorplancad,fan2022cadtransformer,zheng2022gat}, therefore failing to construct a comprehensive representation.
2) \textbf{\textit{Absence of explicit modeling of relationships between primitives of different types.}} Most existing approaches are limited to modeling geometric primitives, thereby neglecting the latent interconnections across different types of primitives\cite{zheng2022gat}. This lack of explicit relational modeling between text and geometry, prevents the network from capturing higher-level structural dependencies, which ultimately constrains the representation capacity and limits overall model performance.

To address above problems, we argue that it is essential to leverage textual annotations in CAD drawings as a vital source of semantic information. By integrating these annotations with geometric primitives, we aim to construct a comprehensive representation that captures both structural and semantic cues. However, achieving this goal faces the following two challenges:
1) \textbf{\textit{Disorder of various textual elements.}} In real-world CAD drawings, text primitives come with irregular ordering and orientations. They may appear at arbitrary locations and directions, and often lack structural regularity\cite{fan2022cadtransformer}. How to effectively utilize these disordered textual primitives to guide representation learning for non-textual primitives remains a critical obstacle to improving training performance.
2) \textbf{\textit{Implicit spatial association among different types of primitives.}} Although different types of primitives coexist within the same scene, the relationships among them are often implicit and unstructured, lacking explicit topological guidance. Capturing these various latent associations across different primitives both spatially and contextually remains a significant challenge for accurate representation learning.

To address the challenges, we propose a panoptic symbol spotting framework for CAD data that incorporates textual semantic information. Specifically, we first decompose CAD drawings into various primitives and construct a graph where text elements are also included as a distinct type of node. Then, a convolution neural network (CNN) is used to extract raster image features, which serve as the initial representation for primitive nodes. Meanwhile, handcrafted edge features are constructed to describe relationships among different symbols. A transformer-based architecture is adopted as the backbone, where a type-aware attention mechanism is introduced to model positional dependencies between different types of symbols. Final results are obtained via a classification head and a clustering head. Therefore, model can jointly learn both structural patterns and semantic associations, enhancing symbol spotting performance. Our main contributions are summarized as follows:
\begin{itemize}
\item \textit{Incorporating textual information into CAD symbol spotting for enhancing representations.} We incorporate textual annotations as a key semantic modality in the CAD symbol spotting task. By combining text with geometric primitives, the model gains a richer understanding of the drawing content, improving representation quality and recognition accuracy in complex scenarios.
\item \textit{Type-aware attention for modeling diverse relationships.} We propose a type-aware attention mechanism to explicitly model the different types of spatial relationships between various primitives. This enhances the model’s ability to understand layout structures and improves its performance on symbol spotting tasks.
\item \textit{Achieved state-of-the-art performance on real-world FloorPlanCAD datasets with text annotations}\cite{fan2021floorplancad}. Our proposed method demonstrates superior performance involving both geometric primitives and textual annotations, validating the practicality and stability of our method to diverse CAD scenarios.
\end{itemize}

\section{PROBLEM FORMALIZATION}
\textbf{Primitives definition.} In a CAD drawing, primitives refer to different types of graphical primitives. In this work, we consider five types of primitives: line, arc, circle, ellipse, and text.
We denote a primitive as $e_{i}$, which is associated with two attributes: $l_{i}$ and $z_{i}$, representing the semantic category and instance index of this primitive $e_{i}$, respectively\cite{fan2021floorplancad}. Primitives corresponding to uncountable stuff are assigned $z_{i}=-1$. Similarly, for primitives those are not part of any specific symbol instance, the instance index is also likewise set to $z_{i}=-1$. A set of primitives sharing the same semantic category and instance index is grouped into a symbol, denoted as $s_{k}=(l_{k},z_{k})$. 
In the symbol spotting task, the goal is to identify the semantic label $\hat{l_{i}}$ and instance index $\hat{z_{i}}$ for each primitives $e_{i}$\cite{fan2022cadtransformer}.

\section{Methodology}
\begin{figure*}[htbp]
\centerline{\includegraphics[width=0.8\textwidth]{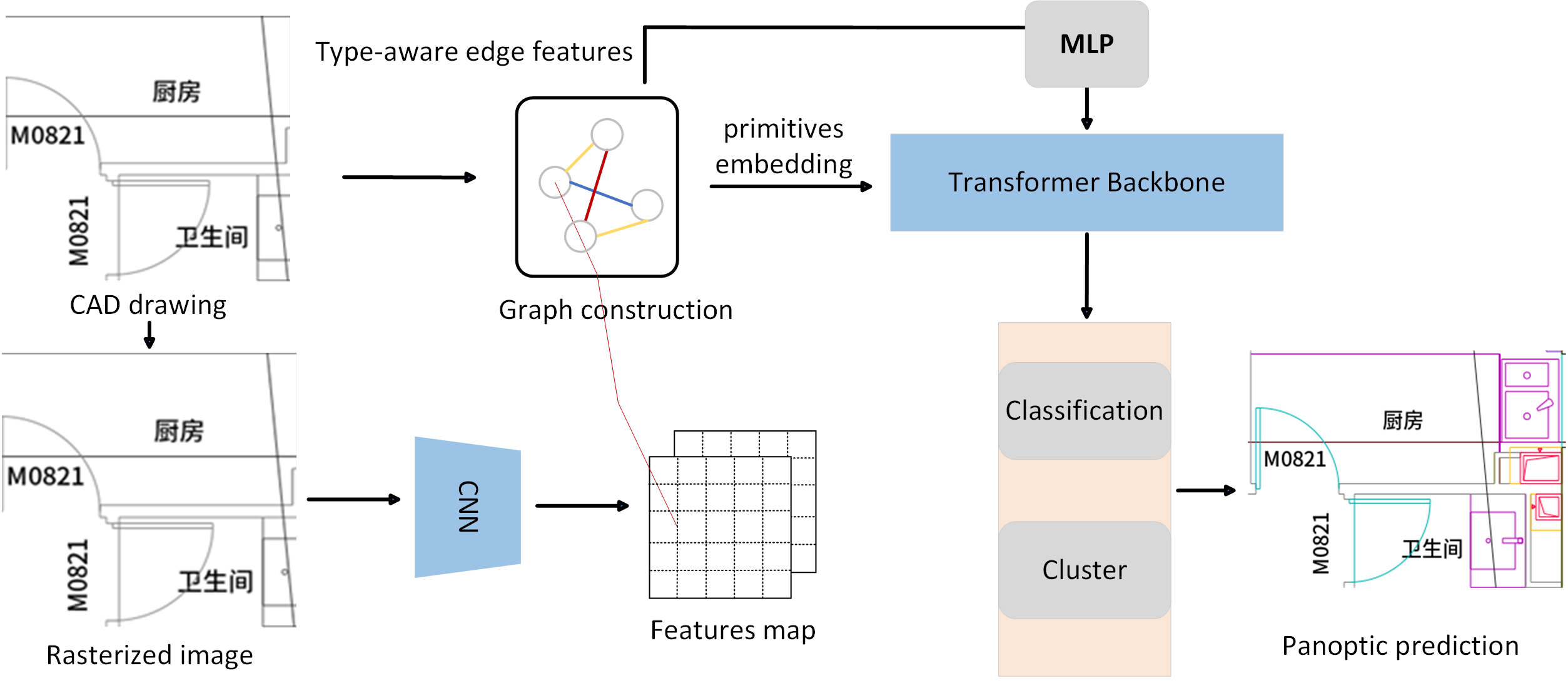}}
\caption{Framework of our method.}
\label{fig2}
\end{figure*}

In this section, we primarily introduce our method. Fig.~\ref{fig2} shows the framework of our model, which consists of several components: \textit{Text Primitives Integration Module} and \textit{Type-aware Attention Mechanism}.

A simplified workflow is as follows.
\textbf{STEP1: Graph construction.} Given an input vectorized CAD drawing $D$, we first decompose it into a set of basic graphical primitives $D=\{p_{k}\}$ (e.g., lines, arcs, circles, ellipse), including text annotations, which are treated as vertexes in a graph. We introduce a text integration module that processes various textual primitives, retaining high-quality annotations with meaningful semantics. These text primitives are incorporated into the graph as a distinct type of primitive node.
\textbf{STEP2: Feature initialization.} A pretrained CNN is used to extract features from the rasterized CAD image\cite{krizhevsky2012imagenet,simonyan2014very}. For one primitive, we sample features from the corresponding spatial location on the feature map as the initial vertex embedding $\mathbf{f}_{0}^{i} \in \mathbf{F}_{0}$. Meanwhile, we manually construct edge features to encode spatial relationships between different types of primitives.
\textbf{STEP3: Feature updating.} We adopt a standard Vision Transformer\cite{dosovitskiy2020image} as the backbone to update node features. To model the spatial dependencies among different primitives, we incorporate a type-aware attention mechanism into the transformer's attention layers.
\textbf{STEP4: Symbol spotting results.} The final primitive representations output by the transformer are passed to a classification head and a clustering head to jointly predict the semantic category and instance grouping of symbols, resulting in the final symbol spotting outcome.

\subsection{Text Primitives Integration Module}
In CAD drawings, text annotations are widely present and carry essential semantic information beyond the geometric layout. These textual elements provide strong guidance for learning discriminative representations of surrounding graphical primitives. 

However, textual annotations in CAD drawings tend to be highly diverse and unstructured. Simply incorporating all text annotations may introduce noise into the model. To address this, \textit{Text Primitives Integration Module} systematically eliminates low-frequency annotations, where the threshold is determined a priori based on corpus statistics, thereby ensuring that only representative and commonly used textual labels contribute to the graph structure. 
%Concretely, annotations that fall below a predefined frequency threshold are excluded from subsequent processing.
%During filtering, we also apply a normalization rule that removes numeric substrings to better merge semantically similar categories.

Given an input vectorized CAD drawing, we adopt primitives as the fundamental processing unit. Due to the intrinsically segmentable format of CAD vectors drawings, it can naturally be decomposed into discrete elements. In addition to standard graphical primitives, we also treat textual annotations as a separate type of primitive and include them in the graph. 

After constructing the graph, we proceed with the feature initialization. Following the approach in\cite{fan2022cadtransformer}, we first rasterize the input CAD vector drawing and extract its visual features map $\mathbf{F}$ using a pretrained CNN. For all types of primitives including text, each of them is projected onto the $\mathbf{F}$ to get the initial feature embedding $\mathbf{f}_{0}^{i} = \varepsilon_{CNN}(\mathbf{F},\mathbf{c}_{i})$, where $\varepsilon_{CNN}$ represents a bilinear interpolation operation applied on $\mathbf{F}$. $\mathbf{c}_{i}$ is the coordinates center of $e_{i}$.
The initialization of edge features is introduced in the next section \ref{edge}.

\subsection{Type-aware Attention Mechanism}\label{edge}  
Although both text and geometric primitives are treated as vertices in the constructed graph, their interrelations are inherently different and often encode implicit spatial structures. To better capture these interactions, we introduce the type-aware edge features encoding to explicitly model the relationships between different nodes. These edge features are then used to enhance the attention mechanism within the Transformer backbone, allowing the model to leverage the latent spatial interrelations and improve the discriminative power of primitive representations.

To represent spatial relationships among primitives, we encode edge features as a combination of two components: A type indicator $\mathbf{t}$ that denotes the category of different node pairs (i.e., graphic primitive-graphic primitive, graphic primitive-text primitive, or text primitive-text primitive). A vector $\mathbf{e} \in \mathbb{R}^{7}$ capturing geometric relations such as relative distance, position, and angle, which is inspired by\cite{zheng2022gat}. The full edge feature is denoted as $\mathbf{E}=(\mathbf{t} \| \mathbf{e}) \in \mathbb{R}^{N \times k \times 8}$, where $N$ represents the number of primitives, and $k$ denotes the number of nearest neighbors selected for each node in order to reduce computational complexity. The neighbor set of the $i$-th primitive is denoted as $\mathcal{N}(i)$.

To incorporate these edge features into the attention process, we propose an type-aware edge guided attention module. All types of primitives are treated as input tokens and their initial feature embeddings $\mathbf{f}_0 \in \mathbb{R}^{N \times d}$ are token representations, where $d$ is the dimensionality of the feature space. We jointly feed $\mathbf{f}_0$ and $\mathbf{E}$ into the Transformer layers as the inputs.

The raw attention score matrix $\mathbf{A}_s$ in the $s$-th stage within the transformer is computed based on the multi-head attention mechanism\cite{vaswani2017attention}. For a given primitives $e_i$, $\mathbf{q}_i^l \in \mathbb{R}^{d}$ denotes the query vector derived from the node feature, and $\mathbf{k}_j^l \in \mathbb{R}^{d}$ be the key vectors corresponding to its $j$-th neighbors where $j \in \mathcal{N}(i)$. 
In the $l$-th attention head, the attention score of primitives $e_i$ attending to its neighbors $e_j$ is defined as:
\begin{equation}
    \alpha_{ij}^l = \frac{\mathbf{q}_i^{l} \cdot \mathbf{k}_j^{l}}{\sqrt{d/h}},
\end{equation}
where $h$ denotes the number of attention heads. The multi-head attention matrix in the $s$-th stage within the transformer backbone $\mathbf{A}_{s} \in \mathbb{R}^{N \times k \times h}$ is then obtained by concatenating attention coefficients across all heads along the last dimension. 
\begin{equation}
    \mathbf{A}_{s} = \text{Concat}(a_{ij}^{1}, a_{ij}^{2}, \dots, a_{ij}^{h}).
\end{equation}

Then, we feed the edge features $\mathbf{E}$ into a multi-layer perceptron (\text{MLP})\cite{tolstikhin2021mlp} to obtain the structural embedding $\mathbf{T}_{s} \in \mathbb{R}^{N \times k \times h}$. This
process can be written as:
\begin{equation}
\begin{aligned}
\mathbf{T}_{s} = \text{MLP}(\mathbf{E}),
\end{aligned}
\end{equation}
% where \text{MLP} composes two \text{Linear} layers with a \text{ReLU}\cite{nair2010rectified} activation.
where \text{MLP} is two linear layers with ReLU\cite{nair2010rectified}.

The structural embedding $\mathbf{T}_{s}$ is then explicitly integrated into the attention score computation as an edge-aware bias term which is similar to the relative position encoding\cite{shaw2018self}. The feature representation of primitives is then iteratively updated as follows:
\begin{equation}
\begin{aligned}
\mathbf{f}_{s} = \text{Softmax}(\mathbf{A}_{s} + \mathbf{T}_{s})\mathbf{f}_{s-1},
\end{aligned}
\end{equation}
where $\mathbf{f}_{s}$ denotes the feature embeddings of all primitives at the stage $s$.
Therefore, the attention mechanism enhances the alignment across different types of relationships between primitives, facilitating more precise symbol spotting by jointly optimizing semantic classification and instance segmentation tasks as follows: 
\begin{equation}
\begin{aligned}
    \mathcal{L} &= \lambda_{sem} \cdot \mathcal{L}_{sem} + \lambda_{ins} \cdot \mathcal{L}_{ins}, \\
\mathcal{L}_{ins} &= \frac{1}{\sum_i m_i} \sum_i \left\| o_i - (c_i - p_i) \right\| \cdot m_i.
\end{aligned}
\end{equation}
Specifically, $\mathcal{L}_{ins}$ follow the definition in \cite{fan2022cadtransformer}, where $m$ is a binary value to mask out primitives of uncountable stuff. $\mathbf{p}_i$ means the coordinates of each primitive $e_i$ and $\mathbf{c}_{i}$ indicates the center coordinates of each instance that $e_{i}$ belongs to. 
%$\mathbf{O} = \text{MLP}(\mathbf{f}^{S}) \in \mathbb{R}^{N \times 2}, \mathbf{o}_i \in \mathbf{O}$ represents the learned offset of each primitives $p_i$, predicted from outputs of Transformer backbone, to the center coordinates belonging instance.
$\mathbf{O} = \text{MLP}(\mathbf{f}_{S}) \in \mathbb{R}^{N \times 2}$, where $\mathbf{o}_i \in \mathbf{O}$ represents the learned offset of each primitives $e_i$ to the belonging instance center, which is predicted from the output $\mathbf{f}_{S}$ of the Transformer.
$\mathcal{L}_{sem}$ is the \text{Cross-entropy} loss computed between the predicted semantic classes and the ground-truth labels.

\section{Experiment}
In this section, extensive experiments are conducted. We report the comparing results and demonstrate the quantitative results of each class using our model. Visualization results of typical cases are illustrated to further validate the effectiveness of our modules and strategy.

\subsection{Experiment Settings}
This section outlines the experimental settings, including the datasets, evaluation metrics, and implementation details.

\textbf{Datasets}. 
We adopt the latest large-scale FloorPlanCAD dataset\cite{fan2021floorplancad} in our experiment, which was released on November 26th, 2021. This dataset includes 15,663 CAD drawings spanning a wide range of real-world architectural contexts, enriched with detailed text annotations. Compared with the initial release and other existing small-scale vector graphics datasets\cite{rusinol2010relational,delalandre2010generation}, this updated version significantly enhances the floor plan dataset in terms of both scale and semantic richness. 
%The dataset has been released in two versions. The initial version released on August 13th, 2021, contains 11,602 vectorized CAD floor plans. However, due to the absence of textual annotations, it does not meet the requirements of multimodal symbol spotting tasks which require the integration of both geometric and textual information. Therefore, this version does not meet the requirements.

%To overcome this limitation, we utilize the updated version released on November 26th, 2021, which significantly expands the original dataset. This version includes 15,663 CAD drawings spanning a wide range of real-world architectural contexts, enriched with detailed text annotations. Compared with the initial release and other existing small-scale vector graphics datasets\cite{rusinol2010relational,delalandre2010generation}, this updated version significantly enhances the floor plan dataset in terms of both scale and semantic richness. 

Specifically, this version of FloorPlanCAD dataset includes 35 object categories with line-wise annotations. The dataset is distinguished between countable “thing” classes (e.g., doors, windows, appliances) and uncountable “stuff” classes (walls, curtain wall, parking spots, row chairs and railing). The thing classes are annotated with both class labels and instance-level index, while the stuff classes are assigned only semantic category labels. Furthermore, each floor plan is spatially divided into regular square blocks of $14\text{m} \times 14\text{m}$ in real-world dimensions, facilitating efficient training and evaluation across dense and large-scale architectural spaces.

\textbf{Metrics}.\label{metrics}
We adopt the evaluation metric proposed in \cite{fan2021floorplancad}, which is specifically tailored for the CAD symbol spotting task. This metric is conceptually similar to the one introduced in \cite{kirillov2019panoptic} for general image panoptic segmentation, but adapted to accommodate the unique characteristics of vectorized CAD drawings. The metric provides a panoptic quality (PQ), defined as the multiplication of two components: recognition quality (RQ) and segmentation quality (SQ). The definitions are as follows:

RQ measures the recognition performance which is equivalent to the widely used F1 score:
\begin{equation}
\begin{aligned}
RQ = \frac{|TP|}{|TP| + \frac{1}{2}|FP| + \frac{1}{2}|FN|}.
\end{aligned}
\end{equation}

A predicted symbol $s_{pred} = (l_{pred},z_{pred})$ is considered a match with a ground truth symbol $s_{gt} = (l_{gt},z_{gt})$ if $l_{pred}=l_{gt}$ and $\text{IoU}(s_{pred}, s_{gt})>0.5$. The intersection over union (IoU) score between two symbols are computed as follows:
\begin{equation}
\begin{aligned}
\text{IoU}(s_{p}, s_{g}) = \frac{\sum_{e_i \in s_p \cap s_g} \log(1 + L(e_i))}{\sum_{e_j \in s_p \cup s_g} \log(1 + L(e_j))}.
\end{aligned}
\end{equation}

SQ measures the segmentation quality by averaging the IoU scores of correctly matched symbol pairs, capturing the geometric alignment quality between predictions and ground truth:
\begin{equation}
\begin{aligned}
SQ = \frac{\sum_{(s_p, s_g) \in TP} \text{IoU}(s_p, s_g)}{|TP|}.
\end{aligned}
\end{equation}

PQ considers both thing and stuff symbols, offering a unified metric for assessing the performance of panoptic symbol spotting approaches:
\begin{equation}
\begin{aligned}
PQ = RQ \times SQ = \frac{\sum_{(s_p, s_g) \in TP} \text{IoU}(s_p, s_g)}{|\text{TP}| + \frac{1}{2}|\text{FP}| + \frac{1}{2}|\text{FN}|}.
\end{aligned}
\end{equation}

\textbf{Implement Details}.
Our experimental settings are as follows. 
In our model training, the architecture is configured with 6 attention heads in the multi-head attention mechanism. For each primitive, the maximum number of associated neighbors is set to $k=16$ and neighbors are selected by k-nearest neighbors (KNN) strategy\cite{cover1967nearest}. 
We adopt the Adam optimizer\cite{kingma2015adam} with parameters $\beta1=0.9$, $\beta2=0.99$, and a learning rate of $2.5\times10^{-5}$. The learning rate is decayed by a factor of 0.5 every 20 epochs. The training is conducted for a total of 50 epochs, and we select the best-performing model based on the validation. $\lambda_{sem}$ and $\lambda_{ins}$ are set to 1 and 0.3, respectively. 
We utilize AM-Softmax\cite{wang2018additive} loss instead of conventional softmax loss.
The above hyperparameters are manually determined based on empirical settings. 
All models, including the baselines, are trained with a fixed batch size of 2. The training is conducted on 2 RTX 3090 GPUs, with one training sample processed per GPU. During training, the total loss gradually decreases as the epochs increases, while performance metrics such as PQ show a consistent upward trend. Both loss and performance metrics tend to stabilize in the later epochs, indicating convergence of the model.

For feature extraction, we employ HRNetV2-W48\cite{sun2019deep} as the pre-trained CNN backbone, which is pre-trained on 1000-class image classification task of ImageNet\cite{russakovsky2015imagenet}.

\subsection{Main Performance}\label{AA}  
In this section, we present the performance of our method and existing baseline on the PQ, RQ, SQ and F1 metrics for panoptic symbol spotting task, as detailed in Table~\ref{mainreresult}. Notably, the best results are highlighted in bold to improve read-ability. The definitions of PQ, RQ and SQ are provided in ~\ref{metrics}. The F1 score refers to the harmonic mean of precision and recall for classification performance evaluation \cite{saito2015precision}.

We first present the comparative performance of our method. As shown in Table~\ref{mainreresult}, our approach achieves superior results across all standard evaluation metrics, including PQ, SQ and RQ, significantly outperforming existing baselines.

Specifically, integrating textual features into CADTransformer results in a notable improvement in PQ from 0.7152 to 0.7352, demonstrating the positive contribution of semantic features in enhancing the overall scene understanding. By explicitly incorporating these features, our model gains better semantic context, leading to more accurate recognition and segmentation of symbol instances.

Furthermore, when spatial relationships among primitives are modeled through our proposed type-aware attention mechanism, PQ is further boosted to approximately 0.7371. This suggests that capturing the relative positioning and interaction patterns between different types of primitives is essential for improving recognition accuracy. Our attention mechanism effectively prioritizes relevant connections across types, enabling the model to learn more discriminative representations.

In addition to PQ, we also observe consistent gains in RQ and SQ comparing to the baseline, indicating that our strategy about text annotations incorporation enhances both symbol recognition accuracy and segmentation precision. The improvement in these metrics reflects better instance-level detection and more accurate boundary prediction, especially for geometrically complex symbols.

\begin{table}[htbp]
\caption{Main Performance on the FloorPlanCAD Dataset}
\centering
{\renewcommand{\arraystretch}{1.4}
\begin{tabular}{|c|cccc|}
\hline
\textbf{Method} & \textbf{PQ} & \textbf{RQ} & \textbf{SQ} & \textbf{F1} \\ \hline
CADTransformer\cite{fan2022cadtransformer}       & 0.7152 & 0.8298 & 0.8619 & 0.7754     \\ \hline
CADTransformer + text& \underline{0.7352} & \textbf{0.8404} & \underline{0.8748} & \underline{0.7834}     \\ \hline
\textbf{Our Method}  & \textbf{0.7371} & \underline{0.8381} & \textbf{0.8794} & \textbf{0.7877}     \\ \hline
\end{tabular}
}
\label{mainreresult}
\end{table}

Additionally, to further investigate the performance across different symbol categories, we provide a class-wise evaluation of PQ, SQ and RQ in Table~\ref{tab2}. We skip some classes with insufficient instance counts.
%as they produce zero values in evaluation and are not statistically meaningful. 
It can be observed that for certain categories the performance shows a slight decline. The textual annotations are highly diverse and sometimes inconsistent across different drawings, which may introduce noise when aligning text with geometric primitives. In categories like bay window where the geometric appearance is complex and annotations are not standardized, the model may be more sensitive to such noise.
Despite these localized declines, our method consistently performs better across the majority of symbol types and the overall trend confirms that our method achieves strong generalization across diverse symbol types. 

\begin{table*}[htbp]
\caption{Class-wise Performance on FloorPlanCAD Dataset}
\label{tab2}
\centering
\renewcommand{\arraystretch}{1.1}
\setlength{\arrayrulewidth}{0.7pt}

% 需要 booktabs + array
\begin{tabularx}{\textwidth}{|l|*{3}{>{\centering\arraybackslash}X}|*{3}{>{\centering\arraybackslash}X}|*{3}{>{\centering\arraybackslash}X}|}
\toprule
\multirow{2}{*}{\textbf{Class}} & \multicolumn{3}{c|}{\textbf{CADTransformer}} & \multicolumn{3}{c|}{\textbf{CADTransformer + text}} & \multicolumn{3}{c|}{\textbf{Our Method}} \\
& PQ & RQ & SQ & PQ & RQ & SQ & PQ & RQ & SQ \\
\midrule
single door      & 0.8032 & 0.8811 & 0.9116 & 0.8185 & 0.8971 & 0.9124 & 0.8180 & 0.8921 & 0.9169 \\
double door      & 0.8491 & 0.9173 & 0.9256 & 0.8612 & 0.9299 & 0.9261 & 0.8600 & 0.9328 & 0.9220 \\
sliding door     & 0.8408 & 0.9118 & 0.9222 & 0.8435 & 0.9098 & 0.9272 & 0.8388 & 0.9095 & 0.9223 \\
window           & 0.6363 & 0.8146 & 0.7811 & 0.7255 & 0.8550 & 0.8486 & 0.7249 & 0.8559 & 0.8469 \\
bay window       & 0.1171 & 0.1739 & 0.6735 & 0.0774 & 0.1266 & 0.6116 & 0.0750 & 0.1075 & 0.6978 \\
blind window     & 0.6518 & 0.7911 & 0.8239 & 0.6938 & 0.8415 & 0.8244 & 0.6964 & 0.8352 & 0.8339 \\
opening symbol   & 0.1642 & 0.2331 & 0.7042 & 0.1227 & 0.1792 & 0.6848 & 0.1764 & 0.2556 & 0.6902 \\
sofa             & 0.5380 & 0.6592 & 0.8162 & 0.5974 & 0.7355 & 0.8122 & 0.6320 & 0.7645 & 0.8266 \\
bed              & 0.6351 & 0.7761 & 0.8184 & 0.5805 & 0.7079 & 0.8200 & 0.6428 & 0.7711 & 0.8336 \\
chair            & 0.6587 & 0.7761 & 0.8487 & 0.6911 & 0.7914 & 0.8732 & 0.6802 & 0.7679 & 0.8858 \\
table            & 0.3931 & 0.5575 & 0.7050 & 0.4633 & 0.5766 & 0.8034 & 0.5291 & 0.6324 & 0.8367 \\
TV cabinet       & 0.6391 & 0.7923 & 0.8066 & 0.7064 & 0.8501 & 0.8310 & 0.7019 & 0.8648 & 0.8117 \\
Wardrobe         & 0.8395 & 0.9372 & 0.8957 & 0.8025 & 0.9329 & 0.8602 & 0.8417 & 0.9525 & 0.8837 \\
cabinet          & 0.5532 & 0.7233 & 0.7648 & 0.5153 & 0.6802 & 0.7575 & 0.5002 & 0.6624 & 0.7552 \\
gas stove        & 0.8562 & 0.9070 & 0.9441 & 0.8630 & 0.9194 & 0.9386 & 0.8705 & 0.9129 & 0.9535 \\
sink             & 0.7674 & 0.8636 & 0.8886 & 0.7716 & 0.8757 & 0.8811 & 0.7705 & 0.8616 & 0.8943 \\
refrigerator     & 0.6509 & 0.7703 & 0.8450 & 0.6181 & 0.7527 & 0.8212 & 0.6015 & 0.7102 & 0.8470 \\
airconditioner   & 0.7494 & 0.8049 & 0.9311 & 0.7031 & 0.7469 & 0.9414 & 0.7182 & 0.7505 & 0.9570 \\
bath             & 0.5436 & 0.6936 & 0.7837 & 0.5065 & 0.6387 & 0.7929 & 0.5497 & 0.6667 & 0.8246 \\
bath tub         & 0.5884 & 0.7484 & 0.7862 & 0.5949 & 0.7409 & 0.8029 & 0.6243 & 0.7712 & 0.8095 \\
washing machine  & 0.7611 & 0.8800 & 0.8649 & 0.7693 & 0.8673 & 0.8869 & 0.7589 & 0.8519 & 0.8909 \\
urinal           & 0.8973 & 0.9549 & 0.9397 & 0.8747 & 0.9469 & 0.9238 & 0.8802 & 0.9505 & 0.9261 \\
squat toilet     & 0.8173 & 0.8971 & 0.9111 & 0.8313 & 0.9038 & 0.9197 & 0.8274 & 0.8954 & 0.9240 \\
toilet           & 0.8316 & 0.9292 & 0.8949 & 0.8541 & 0.9225 & 0.9259 & 0.8657 & 0.9234 & 0.9374 \\
stairs           & 0.7091 & 0.8104 & 0.8750 & 0.7203 & 0.8162 & 0.8825 & 0.7265 & 0.8168 & 0.8894 \\
elevator         & 0.8782 & 0.9304 & 0.9439 & 0.8728 & 0.9424 & 0.9262 & 0.8471 & 0.9234 & 0.9173 \\
escalator        & 0.3794 & 0.5116 & 0.7416 & 0.4025 & 0.5283 & 0.7618 & 0.4706 & 0.6061 & 0.7765 \\
railing          & 0.3555 & 0.4427 & 0.8029 & 0.4250 & 0.5602 & 0.7588 & 0.4257 & 0.5456 & 0.7803 \\
row chairs       & 0.7457 & 0.8571 & 0.8700 & 0.6973 & 0.7692 & 0.9065 & 0.7350 & 0.8118 & 0.9054 \\
parking spot     & 0.6982 & 0.7839 & 0.8907 & 0.6454 & 0.7492 & 0.8614 & 0.6266 & 0.7365 & 0.8507 \\
wall             & 0.6152 & 0.8131 & 0.7567 & 0.6314 & 0.8230 & 0.7673 & 0.6285 & 0.8149 & 0.7712 \\
curtain wall     & 0.4234 & 0.5814 & 0.7283 & 0.4217 & 0.5571 & 0.7568 & 0.4215 & 0.5573 & 0.7563 \\
\midrule
\textbf{total}   & \textbf{0.7152} & \textbf{0.8298} & \textbf{0.8619} & \textbf{0.7352} & \textbf{0.8404} & \textbf{0.8748} & \textbf{0.7371} & \textbf{0.8381} & \textbf{0.8794} \\
\bottomrule
\end{tabularx}
\end{table*}

\subsection{Case Study}
\begin{figure}[htbp]
    \centering
    %\begin{subfigure}[b]{0.45\linewidth}
    \begin{subfigure}[b]{0.42\linewidth}
        \centering
        \includegraphics[width=\textwidth]{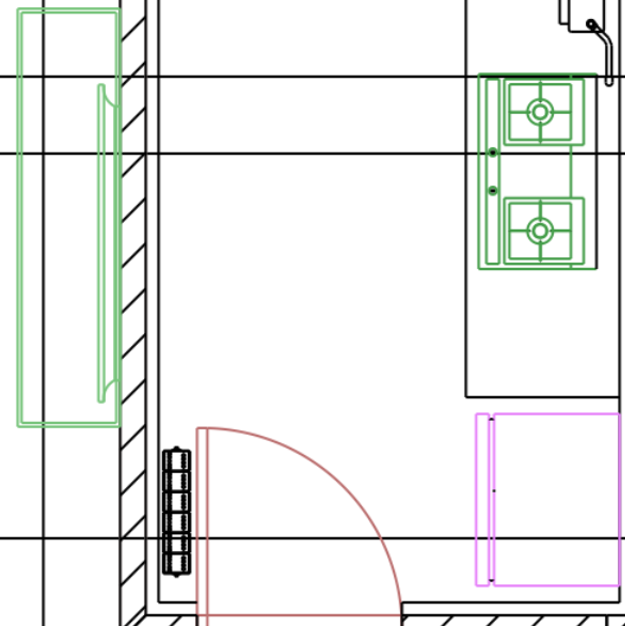}
        \caption{GT}
        \label{fig3:sub-a}
    \end{subfigure}
    \quad % 比 \hfill 紧凑
    \begin{subfigure}[b]{0.42\linewidth}
        \centering
        \includegraphics[width=\textwidth]{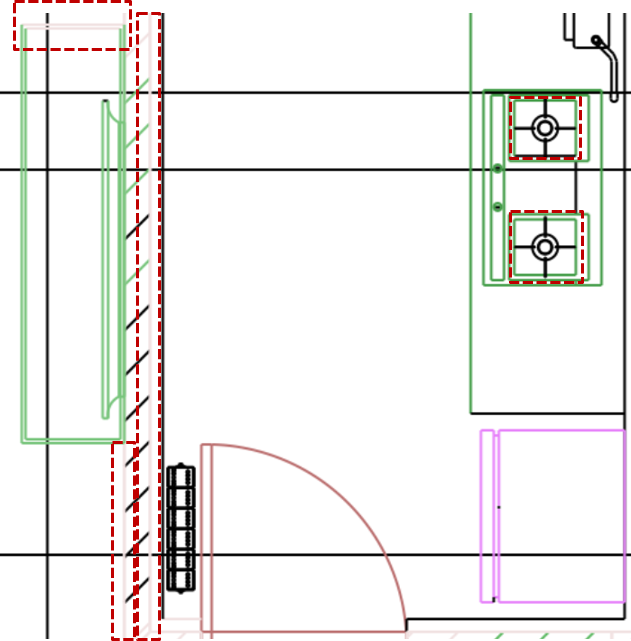}
        \caption{CADTransformer}
        \label{fig3:sub-b}
    \end{subfigure}
    \quad % 比 \hfill 紧凑
    \begin{subfigure}[b]{0.42\linewidth}
        \centering
        \includegraphics[width=\textwidth]{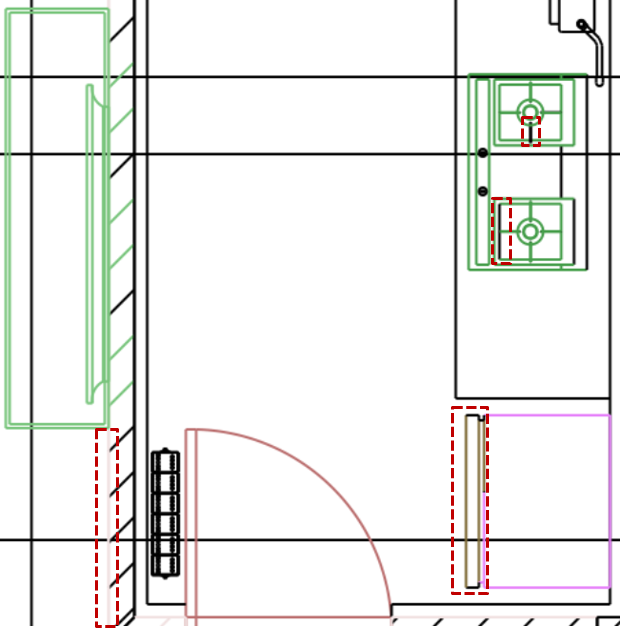}
        \caption{Our method}
        \label{fig3:sub-c}
    \end{subfigure}
    \caption{Visualization results}
    \label{fig3}
\end{figure}

To better demonstrate the effectiveness of our proposed method, we visualize representative case of qualitative evaluation results in Fig.~\ref{fig3}. As shown in the figure, our method outperforms the baselines, particularly in challenging regions that often confuse baseline models.
The red dashed boxes in the figure highlight the regions where the predicted outputs diverge from the ground truth. Compared to CADTransformer, our approach produces fewer misclassifications in these regions, which validate that our model has a stronger ability to handle complex CAD scenarios more robustly.

\section{Related Work}
%Existing approaches for CAD panoptic symbol spotting can be broadly categorized into three main groups: pixel-based, primitive-based and point cloud-based methods. They reflect different perspectives on how CAD drawings should be represented and processed for symbol understanding.
Existing CAD panoptic symbol spotting methods can be grouped into three categories: pixel-based, primitive-based, and point cloud-based, reflecting different perspectives on CAD representation and processing.

%Pixel-based methods treat symbol spotting as a conventional computer vision task, such as object detection or image segmentation task\cite{pang2024pixel, delalandre2010generation, nguyen2009symbol}. While this approach allows the reuse of mature image-based techniques, it often struggles to capture the fine-grained nature of CAD representations. Vector-based elements like lines are reduced to pixels, resulting in a loss of geometric precision. Furthermore, the performance of pixel-based methods tends to degrade when handling large-scale engineering drawings with varying resolutions, scales, and layout complexities. Their high computational cost and reliance on dense raster representations also limit their scalability for industrial applications.
Pixel-based methods treat symbol spotting as an image task, e.g., object detection or image segmentation\cite{pang2024pixel, delalandre2010generation, nguyen2009symbol}. They reuse mature vision techniques but reduce vector elements to pixels, losing geometric precision and struggling with large-scale drawings of varying resolutions. Their reliance on dense rasterization also incurs high computational cost, limiting scalability.

%Primitive-based methods take a more structured approach by directly operating on the geometric primitives inherent in CAD drawings, such as lines and arcs. These methods first decompose drawings into a set of low-level primitives and then leverage either graph-based\cite{fan2021floorplancad,zheng2022gat} or Transformer-based models\cite{fan2022cadtransformer} for representation and recognition. Although these methods preserve the structural information of vector drawings and have achieved impressive results in floor-plan domains, they still face challenges in modeling complex hierarchical relationships among symbols.
Primitive-based methods operate directly on geometric primitives (e.g., lines, arcs), decomposing drawings into low-level elements and using graph-based\cite{fan2021floorplancad,zheng2022gat} or Transformer-based models\cite{fan2022cadtransformer} for representation. These approaches preserve structural information and perform well in floor-plan domains but face difficulties modeling complex hierarchical relationships among symbols.

%Point cloud-based methods represent another emerging direction, where CAD primitives are abstracted as point cloud structures in a high-dimensional space\cite{liu2024symbol}. This allows the application of 3D deep learning techniques such as point cloud analysis and neighborhood aggregation\cite{zhao2021point}. These approaches offer a richer and more flexible representation of geometric information, enabling more robust symbol recognition in cluttered or noisy environments. However, point cloud-based methods tend to focus primarily on geometric features, often neglecting semantic cues embedded in textual annotations.
Point cloud-based methods abstract primitives into point cloud structures in high-dimensional space \cite{liu2024symbol}, enabling 3D learning techniques like point cloud analysis and neighborhood aggregation\cite{zhao2021point}. They capture richer geometric information and are robust to clutter, yet often overlook semantic cues such as textual annotations.

\section{Conclusion}
%This paper proposes a CAD panoptic symbol spotting framework, in which textual annotations are introduced as a key semantic information to enhance representation learning. The design of our framework overcomes the critical challenges in the lack of text elements utilization and spatial relational modeling among different types of primitives. To effectively capture information inherent in CAD drawings, we construct a primitive-level graph that incorporates both geometric and textual primitives. To further enhance the modeling of inter-primitive interactions, we introduce a type-aware attention mechanism that explicitly model the spatial dependencies of different connection types between various primitive. By incorporating text information into the learning process of primitives representation, our method provides a more comprehensive view of the CAD drawing which can improves primitive level spotting accuracy. Extensive experiments on the real-world CAD dataset were performed and the results showed the effectiveness our proposed model. Visualization of qualitative results further confirmed the robustness of our strategy.

This paper proposes a CAD panoptic symbol spotting framework, where textual annotations are introduced as key semantic cues to enhance representation learning. Our design addresses challenges in utilizing text elements and modeling spatial relations among primitives. To capture information inherent in CAD drawings, we construct a primitive-level graph combining geometric and textual primitives. To better model inter-primitive interactions, we introduce a type-aware attention mechanism that explicitly captures spatial dependencies between connection types. Incorporating text into primitive representation learning provides a more comprehensive view of CAD drawings, improving primitive-level spotting accuracy. Extensive experiments on a real-world CAD dataset demonstrate the effectiveness of our model, and qualitative visualizations further verify its robustness.

%\section*{Acknowledgment}

\bibliographystyle{IEEEtran}  
\bibliography{references}     

\begin{thebibliography}{10}
\providecommand{\url}[1]{#1}
\csname url@samestyle\endcsname
\providecommand{\newblock}{\relax}
\providecommand{\bibinfo}[2]{#2}
\providecommand{\BIBentrySTDinterwordspacing}{\spaceskip=0pt\relax}
\providecommand{\BIBentryALTinterwordstretchfactor}{4}
\providecommand{\BIBentryALTinterwordspacing}{\spaceskip=\fontdimen2\font plus
\BIBentryALTinterwordstretchfactor\fontdimen3\font minus \fontdimen4\font\relax}
\providecommand{\BIBforeignlanguage}[2]{{%
\expandafter\ifx\csname l@#1\endcsname\relax
\typeout{** WARNING: IEEEtran.bst: No hyphenation pattern has been}%
\typeout{** loaded for the language `#1'. Using the pattern for}%
\typeout{** the default language instead.}%
\else
\language=\csname l@#1\endcsname
\fi
#2}}
\providecommand{\BIBdecl}{\relax}
\BIBdecl

\bibitem{shivegowda2022review}
M.~D. Shivegowda, P.~Boonyasopon, S.~M. Rangappa, and S.~Siengchin, ``A review on computer-aided design and manufacturing processes in design and architecture,'' \emph{Archives of Computational Methods in Engineering}, vol.~29, no.~6, pp. 3973--3980, 2022.

\bibitem{hirz2017future}
M.~Hirz, P.~Rossbacher, and J.~Gulanov{\'a}, ``Future trends in cad--from the perspective of automotive industry,'' \emph{Computer-Aided Design and Applications}, vol.~14, no.~6, pp. 734--741, 2017.

\bibitem{aouad2013computer}
G.~Aouad, S.~Wu, A.~Lee, and T.~Onyenobi, \emph{Computer aided design guide for architecture, engineering and construction}.\hskip 1em plus 0.5em minus 0.4em\relax Routledge, 2013.

\bibitem{liu2019review}
Z.~Liu, Y.~Lu, and L.~C. Peh, ``A review and scientometric analysis of global building information modeling (bim) research in the architecture, engineering and construction (aec) industry,'' \emph{Buildings}, vol.~9, no.~10, p. 210, 2019.

\bibitem{fan2021floorplancad}
Z.~Fan, L.~Zhu, H.~Li, X.~Chen, S.~Zhu, and P.~Tan, ``Floorplancad: A large-scale cad drawing dataset for panoptic symbol spotting,'' in \emph{Proceedings of the IEEE/CVF international conference on computer vision}, 2021, pp. 10\,128--10\,137.

\bibitem{rezvanifar2019symbol}
A.~Rezvanifar, M.~Cote, and A.~Branzan~Albu, ``Symbol spotting for architectural drawings: state-of-the-art and new industry-driven developments,'' \emph{IPSJ Transactions on Computer Vision and Applications}, vol.~11, no.~1, p.~2, 2019.

\bibitem{rusinol2010symbol}
M.~Rusinol, J.~Llad{\'o}s, and G.~S{\'a}nchez, ``Symbol spotting in vectorized technical drawings through a lookup table of region strings,'' \emph{Pattern Analysis and Applications}, vol.~13, no.~3, pp. 321--331, 2010.

\bibitem{rezvanifar2020symbol}
A.~Rezvanifar, M.~Cote, and A.~B. Albu, ``Symbol spotting on digital architectural floor plans using a deep learning-based framework,'' in \emph{Proceedings of the IEEE/CVF Conference on Computer Vision and Pattern Recognition Workshops}, 2020, pp. 568--569.

\bibitem{nguyen2008symbol}
T.-O. Nguyen, S.~Tabbone, and O.~R. Terrades, ``Symbol descriptor based on shape context and vector model of information retrieval,'' in \emph{2008 The Eighth IAPR International Workshop on Document Analysis Systems}.\hskip 1em plus 0.5em minus 0.4em\relax IEEE, 2008, pp. 191--197.

\bibitem{kirillov2019panoptic}
A.~Kirillov, K.~He, R.~Girshick, C.~Rother, and P.~Doll{\'a}r, ``Panoptic segmentation,'' in \emph{Proceedings of the IEEE/CVF conference on computer vision and pattern recognition}, 2019, pp. 9404--9413.

\bibitem{pang2024pixel}
J.~Pang, Z.~Dong, J.~Deng, M.~Zhu, and Y.~Zhang, ``Pixel-wise symbol spotting via progressive points location for parsing cad images,'' \emph{arXiv preprint arXiv:2404.10985}, 2024.

\bibitem{fan2022cadtransformer}
Z.~Fan, T.~Chen, P.~Wang, and Z.~Wang, ``Cadtransformer: Panoptic symbol spotting transformer for cad drawings,'' in \emph{Proceedings of the IEEE/CVF Conference on Computer Vision and Pattern Recognition}, 2022, pp. 10\,986--10\,996.

\bibitem{zheng2022gat}
Z.~Zheng, J.~Li, L.~Zhu, H.~Li, F.~Petzold, and P.~Tan, ``Gat-cadnet: Graph attention network for panoptic symbol spotting in cad drawings,'' in \emph{Proceedings of the IEEE/CVF conference on computer vision and pattern recognition}, 2022, pp. 11\,747--11\,756.

\bibitem{li2013symbol}
Y.~Li, S.~Lu, and C.~L. Tan, ``Symbol spotting in line drawings through graph matching,'' \emph{Pattern Recognition}, vol.~46, no.~4, pp. 1159--1175, 2013.

\bibitem{krizhevsky2012imagenet}
A.~Krizhevsky, I.~Sutskever, and G.~E. Hinton, ``Imagenet classification with deep convolutional neural networks,'' \emph{Advances in neural information processing systems}, vol.~25, 2012.

\bibitem{simonyan2014very}
K.~Simonyan and A.~Zisserman, ``Very deep convolutional networks for large-scale image recognition,'' \emph{arXiv preprint arXiv:1409.1556}, 2014.

\bibitem{dosovitskiy2020image}
A.~Dosovitskiy, L.~Beyer, A.~Kolesnikov, D.~Weissenborn, X.~Zhai, T.~Unterthiner, M.~Dehghani, M.~Minderer, G.~Heigold, S.~Gelly \emph{et~al.}, ``An image is worth 16x16 words: Transformers for image recognition at scale,'' \emph{arXiv preprint arXiv:2010.11929}, 2020.

\bibitem{vaswani2017attention}
A.~Vaswani, N.~Shazeer, N.~Parmar, J.~Uszkoreit, L.~Jones, A.~N. Gomez, {\L}.~Kaiser, and I.~Polosukhin, ``Attention is all you need,'' \emph{Advances in neural information processing systems}, vol.~30, 2017.

\bibitem{tolstikhin2021mlp}
I.~O. Tolstikhin, N.~Houlsby, A.~Kolesnikov, L.~Beyer, X.~Zhai, T.~Unterthiner, J.~Yung, A.~Steiner, D.~Keysers, J.~Uszkoreit \emph{et~al.}, ``Mlp-mixer: An all-mlp architecture for vision,'' \emph{Advances in neural information processing systems}, vol.~34, pp. 24\,261--24\,272, 2021.

\bibitem{nair2010rectified}
V.~Nair and G.~E. Hinton, ``Rectified linear units improve restricted boltzmann machines,'' in \emph{Proceedings of the 27th international conference on machine learning (ICML-10)}, 2010, pp. 807--814.

\bibitem{shaw2018self}
P.~Shaw, J.~Uszkoreit, and A.~Vaswani, ``Self-attention with relative position representations,'' \emph{arXiv preprint arXiv:1803.02155}, 2018.

\bibitem{rusinol2010relational}
M.~Rusi{\~n}ol, A.~Borr{\`a}s, and J.~Llad{\'o}s, ``Relational indexing of vectorial primitives for symbol spotting in line-drawing images,'' \emph{Pattern Recognition Letters}, vol.~31, no.~3, pp. 188--201, 2010.

\bibitem{delalandre2010generation}
M.~Delalandre, E.~Valveny, T.~Pridmore, and D.~Karatzas, ``Generation of synthetic documents for performance evaluation of symbol recognition \& spotting systems,'' \emph{International Journal on Document Analysis and Recognition (IJDAR)}, vol.~13, no.~3, pp. 187--207, 2010.

\bibitem{cover1967nearest}
T.~Cover and P.~Hart, ``Nearest neighbor pattern classification,'' \emph{IEEE transactions on information theory}, vol.~13, no.~1, pp. 21--27, 1967.

\bibitem{kingma2015adam}
D.~P. Kingma and J.~Ba, ``Adam: A method for stochastic optimization,'' \emph{International Conference on Learning Representations (ICLR)}, 2015.

\bibitem{wang2018additive}
F.~Wang, J.~Cheng, W.~Liu, and H.~Liu, ``Additive margin softmax for face verification,'' \emph{IEEE Signal Processing Letters}, vol.~25, no.~7, pp. 926--930, 2018.

\bibitem{sun2019deep}
K.~Sun, B.~Xiao, D.~Liu, and J.~Wang, ``Deep high-resolution representation learning for human pose estimation,'' in \emph{Proceedings of the IEEE/CVF conference on computer vision and pattern recognition}, 2019, pp. 5693--5703.

\bibitem{russakovsky2015imagenet}
O.~Russakovsky, J.~Deng, H.~Su, J.~Krause, S.~Satheesh, S.~Ma, Z.~Huang, A.~Karpathy, A.~Khosla, M.~Bernstein \emph{et~al.}, ``Imagenet large scale visual recognition challenge,'' \emph{International journal of computer vision}, vol. 115, no.~3, pp. 211--252, 2015.

\bibitem{saito2015precision}
T.~Saito and M.~Rehmsmeier, ``The precision-recall plot is more informative than the roc plot when evaluating binary classifiers on imbalanced datasets,'' \emph{PloS one}, vol.~10, no.~3, p. e0118432, 2015.

\bibitem{nguyen2009symbol}
T.-O. Nguyen, S.~Tabbone, and A.~Boucher, ``A symbol spotting approach based on the vector model and a visual vocabulary,'' in \emph{2009 10th International Conference on Document Analysis and Recognition}.\hskip 1em plus 0.5em minus 0.4em\relax IEEE, 2009, pp. 708--712.

\bibitem{liu2024symbol}
W.~Liu, T.~Yang, Y.~Wang, Q.~Yu, and L.~Zhang, ``Symbol as points: Panoptic symbol spotting via point-based representation,'' \emph{arXiv preprint arXiv:2401.10556}, 2024.

\bibitem{zhao2021point}
H.~Zhao, L.~Jiang, J.~Jia, P.~H. Torr, and V.~Koltun, ``Point transformer,'' in \emph{Proceedings of the IEEE/CVF international conference on computer vision}, 2021, pp. 16\,259--16\,268.

\end{thebibliography}

\end{document}